\documentclass{article}

\usepackage[final]{aloe_2023_neurips}

\usepackage[utf8]{inputenc} 
\usepackage[T1]{fontenc}    
\usepackage{hyperref}       
\usepackage{url}            
\usepackage{booktabs}       
\usepackage{amsfonts}       
\usepackage{nicefrac}       
\usepackage{microtype}      
\usepackage{xcolor}         

\usepackage{graphicx}
\usepackage{caption}
\usepackage{subcaption}

\title{Procedural generation of meta-reinforcement learning tasks}

\author{%
  Thomas Miconi \\
  ML Collective \\
  \texttt{thomas.miconi@gmail.com} \\
}

\begin{document}

\maketitle

\begin{abstract}

Open-endedness stands to benefit from the ability to generate an infinite variety of diverse, challenging environments. One particularly interesting type of challenge is meta-learning (``learning-to-learn''), a hallmark of intelligent behavior. However, the number of meta-learning environments in the literature is limited.
Here we describe a parametrized space for simple meta-reinforcement learning (meta-RL) tasks with arbitrary stimuli. The parametrization allows us to randomly generate an arbitrary number of novel simple meta-learning tasks. The parametrization is expressive enough to include many well-known  meta-RL tasks, such as  bandit problems, the Harlow task, T-mazes, the Daw two-step task and others. Simple extensions allow it to capture tasks based on two-dimensional topological spaces, such as full mazes or find-the-spot domains. We describe a number of randomly generated meta-RL domains of varying complexity and discuss potential issues arising from random generation.
\end{abstract}



\section{Introduction}

A potential key step for open-endedness is the ability to generate a wide variety of diverse, challenging environments. There is of course much ambiguity in what counts as ```challenging''. One possible, reasonably objective criterion would be that each new episode of this environment requires the learning and exploitation of new, unpredictable information. For example, if the environment is ``mazes'', each new maze requires learning anew the reward location and/or maze structure; if the environment is ``bandit tasks', each new instance requires learning and exploiting a new probability distribution over returns, etc. Successfully mastering such an environment would then imply not just learning a fixed set of rules or values, but \emph{learning how to learn} (and exploit) the appropriate information, quickly and efficiently, for each new instance of the domain. This ability, once introduced as ``the acquisition of learning sets'' \cite{harlow1949formation}, is now commonly called ``meta-learning''.

Meta-learning (``learning to learn'') is an important aspect of cognition \citep{harlow1949formation,hattori2023meta} and has elicited much interest in machine learning \citep{thrun98learning,schmidhuber1993reducing,hochreiter2001learning,finn2017model,wang2016learning,duan2016rl2,bengio1991learning,floreano2000evolutionary,ruppin2002evolutionary,soltoggio2008evolutionary,miconi2016backpropagation,miconi2018differentiable}. Recently, meta-reinforcement learning (meta-RL) has been a particularly active area of research \citep{wang2016learning,wang2018prefrontal,duan2016rl2,miconi2018differentiable,laskin2022context}.  Successful Meta-RL agents must develop an in-built RL algorithm specifically adapted to the environment, allowing for fast and efficient acquisition of each new individual task from the domain. Recent research has often focused on increasing the \emph{complexity} of  meta-RL environments, making use of combinatorial rule spaces, physical grounding, long horizons, etc. \cite{wang2021alchemy,team2023human}; or alternatively, making the meta-domain adversarial by biasing it towards individual tasks difficult for the agent \cite{jackson2023discovering}. 

On the other hand the total \emph{number} of meta-learning domains in the literature has remained relatively limited.  We note that certain potential applications (such as robustly evaluating meta-learning algorithms, or perhaps even training them in a meta-meta-learning process) require a large number of different meta-learning tasks. More fundamentally, the limited number of known meta-learning tasks reduces the potential of meta-learning environments for open-endedness.

One way to greatly expand the number of available meta-tasks is random procedural generation. Randomly generating meta-tasks requires a \emph{parametrized space} of meta-tasks, which can be explored by varying the values of the parameters. Here we describe one possible parametrization for meta-RL tasks, which is both simple and expressive enough to capture commonly used  meta-RL tasks from the literature.

We emphasize that we are not proposing a new meta-RL domain, from which one can generate complex individual RL tasks (as in \cite{wang2021alchemy,team2023human}); rather, we are proposing a method for generating an arbitrary number of simple, but full-fledged \emph{meta}-RL tasks. The fact that the generated domains are truly meta-RL tasks is demonstrated by construction, by showing that the system can generate candidate domains that are commonly used in the literature and universally regarded as true meta-tasks (such as the Harlow meta-task, bandit problems, simple mazes, ``dark room'', etc.) 

All code for this paper is available at 
{\small \texttt{github.com/ThomasMiconi/Meta-Task-Generator}}

\section{A parametrized space of meta-learning tasks}

\begin{figure}
    \centering
    \includegraphics[scale=.25]{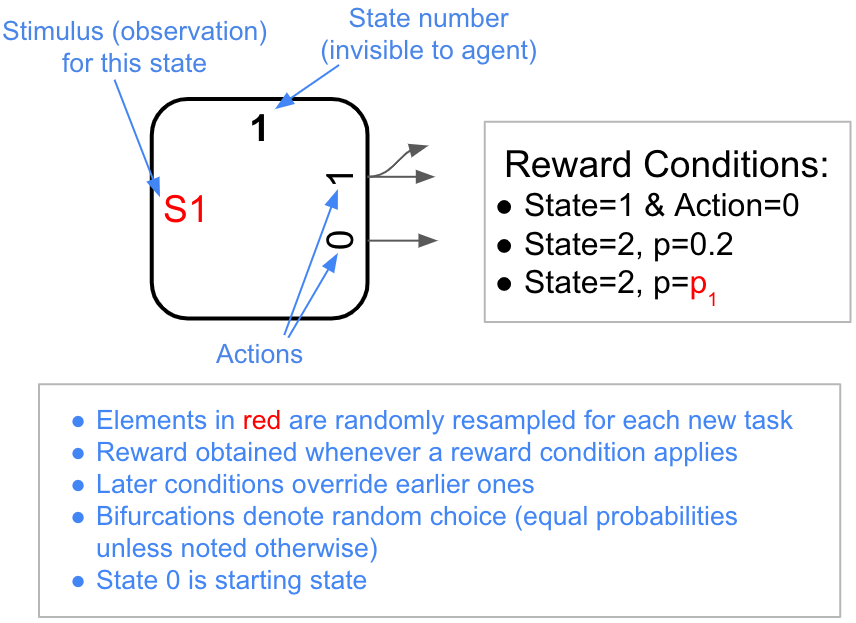}
    \caption{Conventions for the graphical description of meta-learning tasks. Each meta-task is specified as a Partially Observable Markov Decision Process (POMDP). Each rounded square represents a state of the POMDP. State ID is not directly accessible to the agent; only the observation/stimulus for this state (indicated on the left side of the square), which may be null, is provided. Possible actions are indicated on the right side of the square. Arrows indicate future possible state for each action; bifurcated arrows indicate random choice between two possible next states, with equal probabilities unless stated otherwise. Reward is delivered to the agent whenever one of the reward conditions is met; if multiple conditions are met, the applicable reward condition furthest down the list is used. Elements in {\color{red}red} are variables, meaning that they are randomly resampled for each new task generated from the meta-specification.}
    \label{fig:conventions}
\end{figure}

\subsection{Meta-RL tasks as POMDPs with variable quantities}

We describe individual RL tasks as partially-observable Markov decision processes (POMDP).  In POMDPs, unlike standard Markov decision processes, the agent is not explicitly informed of which state it is in. Instead, each state may provide a certain stimulus (``observation'') to the agent, which in general will not uniquely identify the state, but can be used (in addition to other context, past or present) to infer the state. 

Accordingly, in our parametrization, an individual RL task is described as a set of states  $S$, each of which may provide a  certain stimulus (observation) $o_s$ to the agent. In each state $s \in S$, a certain set of actions $a_i \in A$ are available. Each action causes a transition to a next state $s'=s_{t+1} \in S$ with a certain probability determined by the transition function $T(s_t, a_t)$. In addition,  each action may provide a reward, as determined by the (possibly probabilistic) reward function $r(s_t, a_t, s_{t+1})$.

In our formulation, the reward function is described as a set of rules of the following form: ``if starting from state $s$, taking action $a$, and ending up in new state $s'$, receive reward $r$ with probability $p$''.  Importantly, $s$, $a$ and $s'$ may be left unspecified  (``don't care''), allowing the rule to be triggered for any value of the unspecified variables (though at least one of them must be specified for each rule). 

Furthermore, the rules for a given meta-task are provided in an ordered sequence, such  that later rules in the sequence may override earlier one. For example, if a rule early in the sequence specifies that taking any action in state 2 results in reward 1 with probability 1, while  a later rule specifies that taking action 0 in state 2 results in reward 2 with probability 0.5, the former rule will be applied if the agent (in state 2) takes any action  other than 0, but if it takes action 0 then the latter rule will apply.

A POMDP with fixed rules, as described above, describes a single reinforcement learning task. To turn this into a meta-learning task specification, we must allow various quantities in the POMDP to be specified by \emph{variables}. These variables are then assigned randomly resampled values (from a pre-specified domain) for each new individual task. Variable quantities may include the probabilities for each reward rule, the state or action in which it applies, and the stimulus (observation) provided at any given state (other quantities, such as transition probabilities, might be specified as variables, but this will not be explored here).  Only some quantities  will be variable across  individual tasks, while others  will be identical across all tasks and thus be part of the meta-task specification. Importantly, a given variable may be used multiple times in a specification, and each occurrence of the variable is assigned the same value for each new task, allowing for coherent variation across task instances. 

\subsection{Basic Examples}

\begin{figure}
    \centering
    \includegraphics[scale=.4]{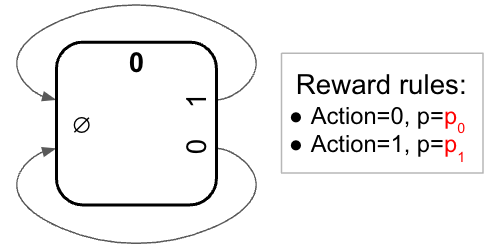}
    \caption{A simple two-arm bandit task. Probabilities  of reward for each arm ($p_0$ and $p_0$) are randomly resampled for each new task instance, as indicated by red coloring.}
    \label{fig:bandit}
\end{figure}


\subsubsection*{Conventions}

In all examples in this paper, the reward value for any triggered rule is always 1, with probability 1 unless specified otherwise. There are two possible actions (``0''  and ``1'') for each state. Reward rules do  not make use of the post-action state $s'$. Also, state 0 is always the starting state for each new episode, and episodes are ended after a fixed number of time steps (removing the need to specify a termination state). This convention simplifies descriptions while still allowing many meta-tasks to be described. 

For graphical representations (Figure \ref{fig:conventions}), each rounded square represents a state of the POMDP. Each bifurcating arrow indicates a random choice between two possible outcomes (next states), with  equal probability unless stated otherwise. Furthermore, if only one arrow emerges from a given state (as opposed to one separate arrow  for each action), it means the next-state transition probabilities for both actions are identical; however, the reward for both actions may still be different. See Figure \ref{fig:conventions} for a summary.   Variables are indicated in red, while black lettering indicates values that are fixed across all individual tasks.

\subsubsection*{Basic two-arm bandit}

A very simple example of a meta-task specification is the simple two-arm bandit problem, as shown in Figure \ref{fig:bandit}.  There is only one state, with no stimulus (observation is empty), and both actions loop back to the same state. The reward probabilities depend on which action is taken, and are randomly resampled (from a uniform distribution over $(0,1)$) for each new task.

\begin{figure}
    \centering
    \includegraphics[scale=.3]{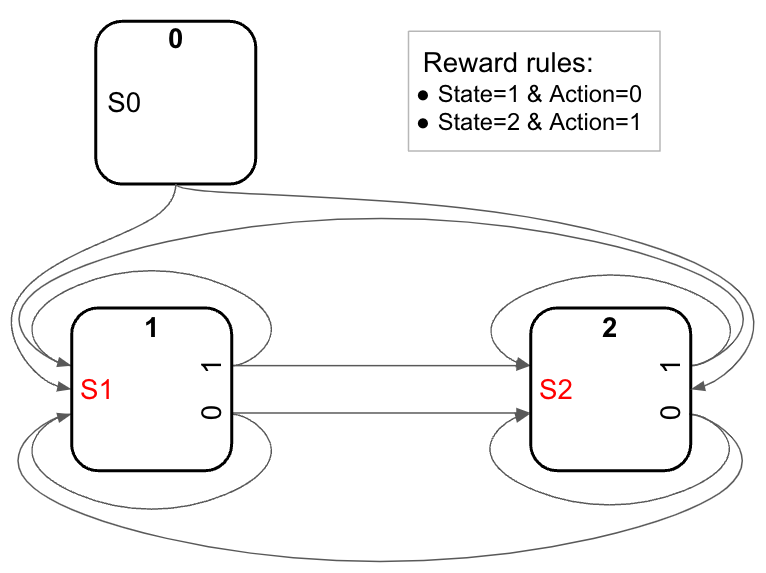}
    \caption{Harlow task with sequential, random-order object presentation. State 1 (resp. 2) is the presentation of the rewarded (resp. unrewarded) stimulus. Action 0 means ``select'', while action 1 means ``ignore''. See text for details.} 
    
    \label{fig:harlow}
\end{figure}

\subsubsection*{Harlow task with sequential object presentation}

A slightly  more elaborate example is the Harlow meta-task \cite{harlow1949formation} with sequential object presentation, as described by \citet{wang2016learning} (end of Section 3.2.2) and \citet{goudar2023schema}. In each episode of  this task, two objects (the ``rewarded'' object and the ``non-rewarded'' object)  are repeatedly presented in random order. The agent receives a reward for picking the rewarded object, and for ignoring the non-rewarded object, when either is presented. Importantly, two completely different objects are used for each new episode (that is, for each  new instance of the meta-task), and each time the agent must find out which of them is the rewarded one from rewards alone. 

This task is described in Figure \ref{fig:harlow}. Actions 0 and 1 correspond to ``select''  and ``ignore'', respectively. State 1 (resp. 2) represents the presentation of the rewarded (resp. unrewarded) object. The appearance of the object is  represented by the  stimuli (observations) S1 and S2  associated with each state; as indicated by the red coloring, these are variables, and thus are randomly regenerated for each new individual task,  representing the two novel objects of each episode.

\subsection{Special states}

\begin{figure}
    \centering
    \includegraphics[scale=.3]{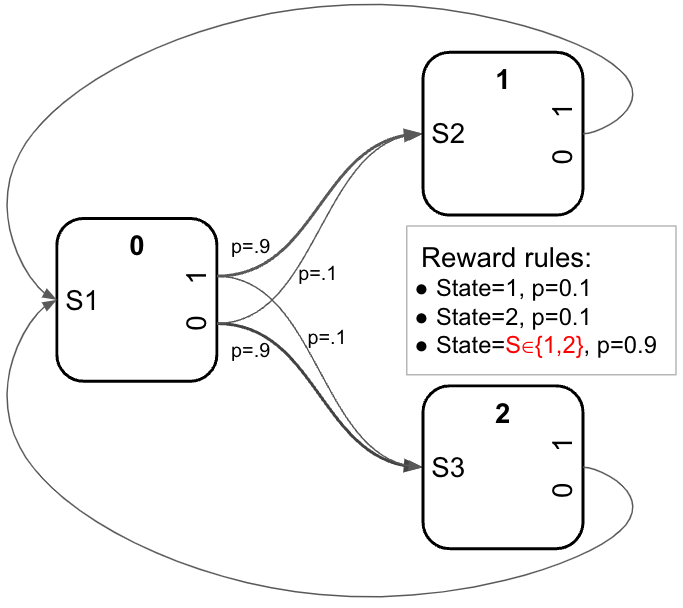}
    \caption{The Daw two-step task, in the version described by \citet{wang2016learning}.}
    \label{fig:twostep}
\end{figure}

Some  meta-tasks may require that one specific state be in some sense ``special'',  in a way that affects various aspects of the environment consistently. Which state is to be the ``special'' one varies from one individual task to the next. For example, in a typical maze meta-task, the reward location changes across instances of the task (i.e. for each individual maze), affecting the reward function. We model this by allowing a meta-task to declare one or more ``special state'' variables, that can be used in the reward rules instead of definite state numbers. Then, when generating each new individual task, we randomly assign a specific state number to each special-state variable, and replace each occurrence of this special-state variable in the rules with its assigned actual state number. Importantly, each occurrence of a given state variable is replaced by the same state number, and different state variables are replaced with different states, which allows for coordinated changes across the system.

As an example, consider Figure \ref{fig:twostep}, which describes the Daw two-step task (in the version implemented by \citet{wang2016learning}). In the initial state, the agent  must choose one of two actions, each of which can lead to either of two second-step states, with opposite probabilities (action 0 is much more likely to lead to state 1, while action 1 is much more likely to lead to state 2 - but both have nonzero probability to lead to both state 1 and 2); these probabilities are fixed for the meta-task across all task instances. Both  second-step states probabilistically provide reward, but one  provides much more reward than the other. Which of  the two states is more rewarded varies randomly from one task to the next and must be inferred by integrating  actions, stimuli and rewards. 

Importantly, a competent learner should take task structure into account: in some cases, receiving a reward after a certain choice should actually \emph{reduce} the probability of taking this particular choice in the future, if the choice was followed by a rare transition to the less frequent next state for this choice (since it suggests that this less frequent state is likely to provide reward, and is more reliably accessed by taking the other choice).

This meta-task is represented very simply with special-state variables.  The reward function specifies that  either of the two second-step states (1 and 2) provides reward with some low probability, and then  the special state variable $S$ 
 (which is randomly resampled from the set ${1,2}$  for each new task instance) indicates which second-step state provides reward with high  probability. Since latter rules override earlier ones, the latter will apply  whenever the agent is in the corresponding state.

\subsection{Statefulness: flag variables}

In POMDPs, statefulness is encoded by the current state (though it may not be directly perceptible to the agent). However, in addition, it may be convenient to allow for additional statefulness, in which some trace of previous agent actions may be stored in readily accessible variables separately from the actual POMDP state. For example, in a key-door task (e.g.  \citep{laskin2022context}), the agent receives reward on reaching the ``door'' location, but only if it has picked the ``key'' beforehand. Such a task can be implemented in a strict POMDP by by duplicating each state according to whether or not the key has been picked. However, it is much simpler to explicitly store whether the agent has picked the  key in an accessible variable, on which transition and reward functions may be conditioned.

We implement a simple type of statefulness in the form of ``flag'' variables. The meta-task may specify  one or  more ``flag'' variables, which are set to 0 at initialization, and can be set to 1 under certain conditions. These conditions are specified as rules conditioned  on $s_t$ and $a_t$,  much like rewards. Furthermore, rewards themselves can now be conditioned on the current value of flag variables:  reward rules include a ``flag'' field  which specifies a necessary value of the  flag for the rule to apply (possibly  including ``don't care'').

\begin{figure*}
    \centering
    \includegraphics[scale=.3]{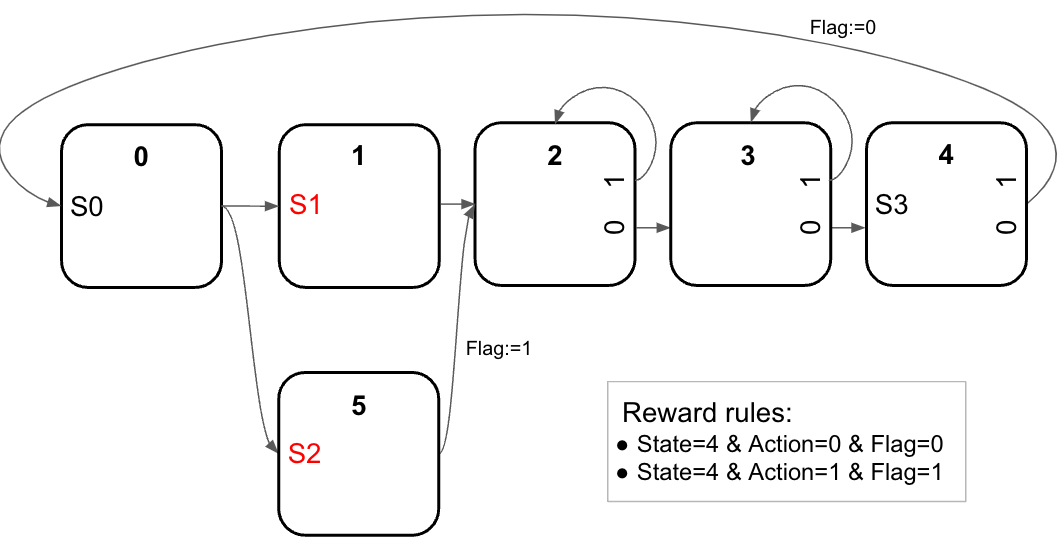}
    \caption{A  simple T-maze task. Either stimulus instructs a different choice in the final state, and sets the flag variable accordingly. Reward delivery after the final state depends on action taken and current flag value.}
    \label{fig:tmaze}
\end{figure*}

As an example, consider the  simple T-maze meta-task  described in  Figure \ref{fig:tmaze}. In the first state, one of two possible stimuli is shown, each of which encodes a direction (``left'' or  ``right'')  to be followed at the end of the  maze. The  agent must then go through a number of intermediate  states (actions 0 and 1 encoding ``stay put'' and ``forward'', respectively - optimal behavior is to proceed forward without delaying).  The agent then enters the ``T-junction''  state, at which point actions 0 and 1 encode ``left'' and ``right''  respectively, and  must choose the one instructed by the initial  stimulus. Both actions transition back to the initial state for a new trial. The content of the ``left''  and ``right'' stimuli are randomly resampled for each new instance of the task.

Without flag variables, describing this task requires duplicating each state depending on which stimulus was shown in state 0. By contrast, with flag variables, the right-coding stimulus sets the flag to 1. The flag is then used at the T-junction state, together with which action was taken, to determine reward for this trial. This considerably simplifies the description of the meta-task. In turn, this simplification implies that a much larger set of meta-tasks can be covered, at the cost of  minimal additional information.

Optionally, the meta-task may also specify that all flags should be reset to 0 every time the agent re-enters  the initial state. This introduces a multi-trial structure to the episode, with each ``trial'' starting every time the system re-enters the initial state, with flags reset at 0.

\subsection{Implementing novel meta-tasks}

As a further illustration, we show that the framework can describe novel meta-RL tasks not previously reported in the meta-learning literature, though  reminiscent of animal training experiments. We describe a stay/switch bandit task (in which the agent must decide to stay with the current arm or switch to a different arm), and a familiarity detection task. See Section \ref{sec:noveltasks} for details.

\section{Random generation of novel meta-RL tasks}


\begin{figure}
    \centering

    \begin{subfigure}[b]{0.45\textwidth}
    \includegraphics[scale=.25]{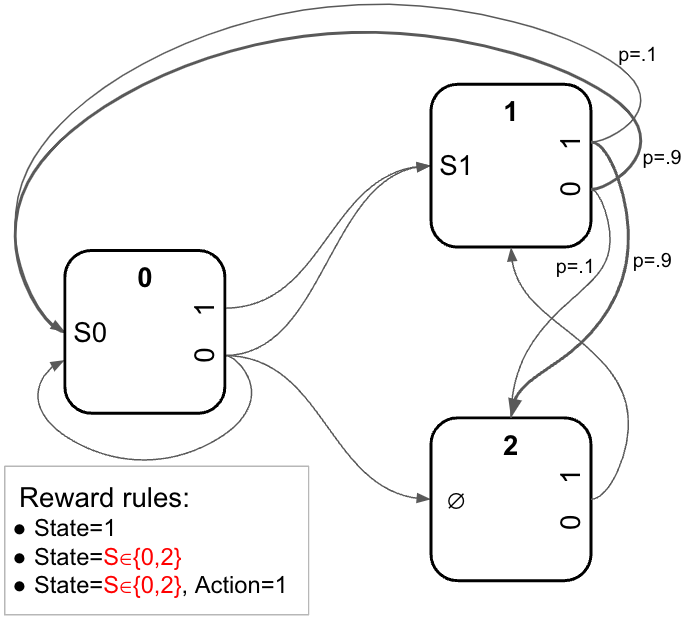}
    \caption{First}
    \label{fig:task1}
    \end{subfigure}
    \begin{subfigure}[b]{0.45\textwidth}
    \includegraphics[scale=.25]{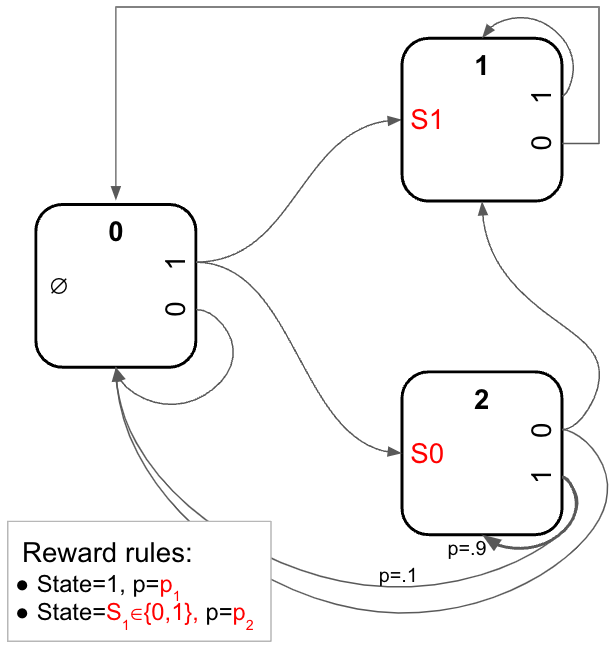}
    \caption{Second}
    \label{fig:task2}
    \end{subfigure}

    \caption{First and second randomly generated meta-task.}
\end{figure}

We now describe several meta-RL tasks randomly generated from the space described above.  The meta-tasks in this section were selected for potential interest and are shown just as they were generated, without any modification.  All code is available at
{\small \texttt{github.com/ThomasMiconi/Meta-Task-Generator}}.


\subsubsection*{First meta-task}

In Figure \ref{fig:task1} we show a simple meta-task with three states, including initial state 0. In state 0, action 0  always lead to state 1, while action 1 randomly leads to any of the three states. From state 2, the only possible transition  is to state 1. The relevant choice in the task is in state 1, where actions 0 and 1 both lead to either state 0 or 2, but with opposite probabilities. Furthermore, from the list of reward rules, we see that state 1 is  always rewarded, and in addition the special state (which can be either 0 or 2 for any  given task instance) is always rewarded (note that the last rule is redundant given the second rule). State 0 and 1 are given different stimuli  (which are fixed across all instances),  while state 2 provides no stimulus; thus all states are identifiable in zero-shot given enough meta-training.

The optimal strategy depends on whether the special state for the current task is 0 or 2. In state 0, the best strategy is always to take action 0. If the special (rewarded) state is 0, the optimal strategy in state 1 is to take action 1, which maximizes occupancy of rewarded states 0 and 1. If the special state is 2, the optimal strategy in state  1 is to take action 0, which maximizes occupancy of rewarded states 0 and 2. Since the special state is not known  to the agent a priori, which of these conditions hold must be determined by exploration for each new task. Note that in state 2, the action taken  is irrelevant. Thus, the task essentially amounts to a bandit task on the two actions of state 1.


\subsubsection*{Second meta-task}

Figure \ref{fig:task2} shows a different randomly  generated meta-task. State 1 is  always rewarded, while state 0 may be rewarded if it  is the special state (as randomly sampled for each new instance of the  meta-task). Furthermore, the reward probabilities are also randomly resampled for each new task. If the special state is 1 (meaning that only state 1 provide reward), then the optimal policy is to get to state 1 and stay there (by action 1). If the special state is 0, and if the reward for state 0 is much higher than that for 1, the optimal policy is to simply stay in state 0 by always invoking action 0 in state 0. Again, to assess which of these two conditions hold, the agent must explore the environment and adequately integrate received rewards. It must then decide which state to come back to (0 or 1), successfully navigate to this state and stay there. Building these exploration and exploitation strategies is a non-trivial challenge for the outer loop of the meta-learning process.

\subsubsection*{Third meta-task: spontaneous key-door task}

Finally, we show a slightly  more elaborate meta-task  that involves both flags and special states. Coincidentally, this randomly generated meta-task implements a crude ``key-door'' environment. The agent collects reward any time it enters state 2 (``door''), but only if the flag is  set to 1 (the ``key''  has been picked). The flag is set to 1 when the agent enters a special (``key'') state, which can be any of the non-initial states  - including state 2 itself. Note that, coincidentally, the transition matrix indicates that after entering state 2 the agent ``teleports''  to a random state other than 2, reminiscent of maze tasks where the agent is teleported on reaching the reward location  \citep{miconi2018differentiable}. Two stimuli are variable, but each state is clearly identifiable for a trained agent (notice that variable state 0 is always the first one seen for each episode).

\begin{figure}
    \centering
    \includegraphics[scale=.3]{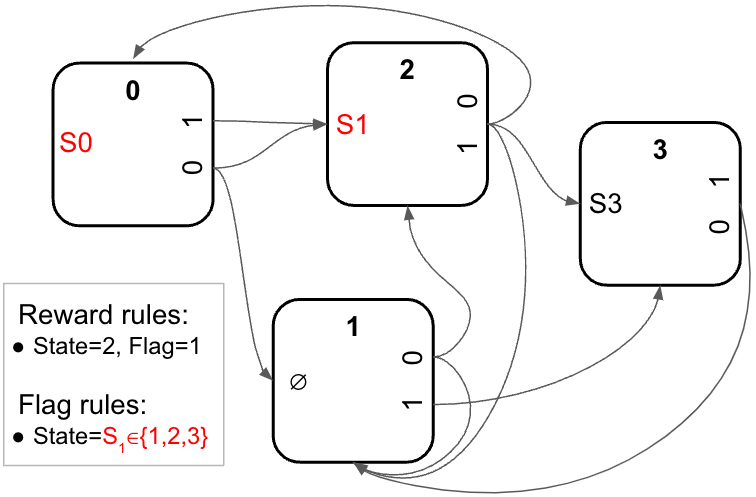}
    \caption{Third randomly generated meta-task. Note that this meta-task coincidentally implements a simple type of ``key-door'' task, with the ``key'' in special variable state $S_1$ (randomly assigned as state 1, 2 or 3 for each new task) and the ``door'' in state 2.}
    \label{fig:task3}
\end{figure}

\section{Extension: tasks with topological structure}

In theory, the above framework can include tasks with two-dimensional (2D) topological environments, such as mazes, Dark Room (``find-the-spot''), key-door, etc. However, such task possess considerable regularities that would be unlikely  to occur by chance in random generation. Fortunately, this structure  can easily be incorporated  in the framework as a potential extension.

To accommodate 2D structure, the states should be conceptually arranged in a 2D grid. Each state must possess enough actions to represent all possible movements within a neighborhood (e.g. 4 actions for up, down, left and right). Furthermore,  the transition probabilities should  reflect the topology, connecting neighboring states through the appropriate actions. Finally, in addition to whatever stimuli / observations the task requires, each state $s$ may provide its $x$ and $y$ coordinates in the overall grid as part of its associated observation $o_s$.

\begin{figure}
    \centering
    \includegraphics[scale=.25]{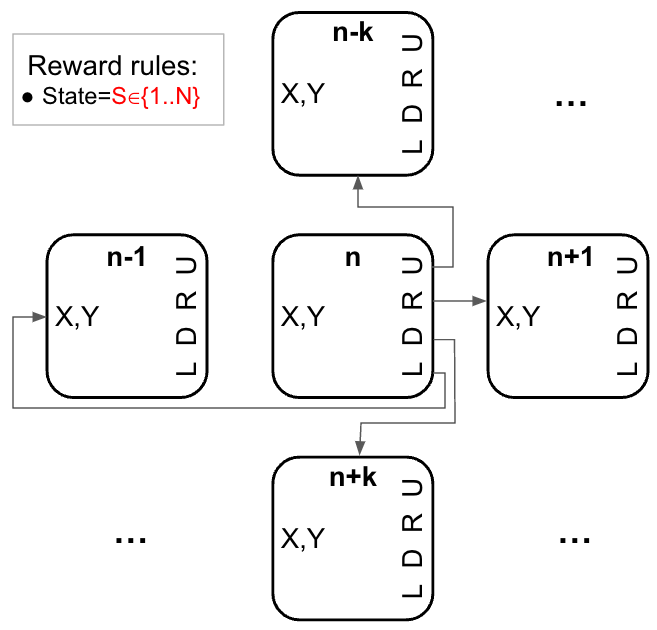}
    \caption{The Dark Room task described in \citet{laskin2022context}, as a simple example of  tasks with 2D topological structure. One location, randomly resampled for each new task instance, provides reward. Each state provides its X and Y coordinates as observation to the agent.}
    \label{fig:darkroom}
\end{figure}

In Figure \ref{fig:darkroom}, we represent the simple ``Dark Room'' experiment described in \citet{laskin2022context}. The goal of the task is simply to find the rewarded spot in  a two-dimensional environment, with no other stimulus than the current $(x,y)$ position (it may be seen as a discrete variant of the Morris maze).  We define the rewarded spot as a special state, to be randomly chosen (from the entire set of states) for each new task instance. Again, the framework can easily be extended to allow special states to influence the transition matrix in addition to rewards, though we do not  explore this further in this paper.

\section{Machine-readable specifications of meta-tasks}

The formal space allows for the description of meta-tasks in machine-readable format. In Appendix \ref{sec:json}, we provide an example of a machine-readable (JSON) specification for the third randomly generated meta-task above (see Figure \ref{fig:json}). This automatically-generated JSON description fully specifies the meta-task, allowing for automated sampling of individual task from the specification.

\section{Potential issues with random generation of meta-RL tasks}

\label{sec:isooptimal}

The space described above includes meta-tasks that are not all  of equal interest. 
For example, if no variable element is included, then obviously all tasks from the  specification will be identical. Even if some variable elements are specified, the resulting set of tasks may turn out  to be equivalent, in the sense that any given policy would obtain the exact same returns on all tasks from the specification. 

A less obvious risk is that the individual tasks would not be equivalent, but \emph{iso-optimal} - that is, some meta-tasks might happen to have the same \emph{optimal} policy for all individual task instances. This may occur even if the actual returns of this single optimal policy are different across task instances,  as long as it is still the optimal policy for each new task instance. These single-optimum meta-tasks are clearly of reduced interest, because they  remove the need for learning-to-learn : once the optimal strategy is discovered, any  additional task instances can be solved by the exact same policy without any further in-task learning. 

Fortunately, this risk can be addressed by filtering. Because the individual tasks are simple POMDPs, they can be solved exactly in reasonable time by known algorithms. Therefore, a filtering process for iso-optimal meta-tasks would be to sample $N$ individual tasks from the meta-task specification, solve them (producing an optimal policy $\mathbf{\pi}_n$ for each task $n$), and then test the performance $R_{k}(\mathbf{\pi}_{j \neq k})$ of each optimal policy $j \neq k$ on each task $k$. If $R_{k}(\mathbf{\pi}_{j \neq k}) = R_{k}(\mathbf{\pi}_{k})$ for all tasks $k, j \neq k$, then the optimal policy for any task $j \neq k$ is also an optimal policy for each task $k$, and the meta-task can be flagged as having a single optimal strategy.

\section{Conclusion}

We have introduced a formal parametrization for meta-RL tasks, which can be used to randomly generate a wide variety of meta-tasks. Although the specification is conceptually and computationally simple (POMDPs with variable quantities), it is expressive enough to cover commonly used meta-tasks from the literature, such as the Harlow (meta-)task, bandit problems, the Daw 2-step meta-task, simple mazes etc. Importantly, this ability confirms that the problems generated by sampling from this parametrization are truly meta-RL tasks, rather than just basic RL tasks sampled from a single, complex meta-RL domain (as in \cite{wang2021alchemy,team2023human}).

The ability to generate arbitrary numbers of different meta-RL tasks (provided that care is taken in ensuring this difference, see Section \ref{sec:isooptimal}) has various applications. One possible application could be to robustly evaluate and compare various meta-learning algorithms. Another would be to train systems capable of autonomous meta-learning, that is, self-contained systems that can learn to learn when faced with a novel domain never seen before in training or evolution, much as humans and animal subjects do in behavioral experiments \cite{harlow1949formation,morcos2016history}.

We note that the specification offers multiple parameters that directly influence the complexity of the resulting meta-task, such as the number of states, the number of variable quantities, the dimensionality of the stimuli, etc. As such, it seems particularly amenable to  automatic curricula, generating tasks of variable and controllable complexity to accompany progress in the (meta-meta-)learning process.

\bibliography{biblio}
\bibliographystyle{apalike}

\appendix

\section*{Supplementary Materials}

\renewcommand\thefigure{S\arabic{figure}}    
\setcounter{figure}{0}

\section{Description of novel meta-learning tasks}

\label{sec:noveltasks}

Here we show that the framework can describe novel meta-RL tasks not previously reported in the meta-learning literature, though  reminiscent of animal training experiments. 

\subsubsection*{Stay/switch bandit task}

\begin{figure}
    \centering
    \includegraphics[scale=.3]{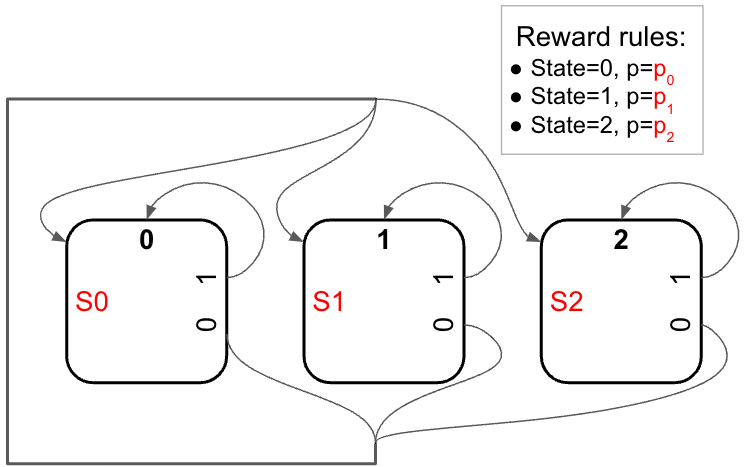}
    \caption{Stay/switch bandit task. Action 0 is ``switch'', action 1 is ``stay''. The agent must repeatedly explore the various states to estimate their respective reward probabilities.}
    \label{fig:stayswitch}
\end{figure}

We describe a stay/switch bandit task. Each of three stimuli is associated with a different probability  of reward. On perceiving a given stimulus,  the agent can choose to ``stay'', and experience this stimulus again (with its associated reward probability) in the next time step; or ``switch'' to a  randomly chosen stimulus  among the three. The three stimuli, and their associated reward probabilities, are randomly resampled for each new task instance. Solving each instance of this meta-task requires identifying the  stimulus with the highest reward probability, which requires exploration (``switches'') and integration of rewards (``stays'').  As shown in Figure \ref{fig:stayswitch}, this meta-task is easily described in our formalism.

\subsubsection*{Familiarity detection}

\begin{figure*}
    \centering
    \includegraphics[scale=.3]{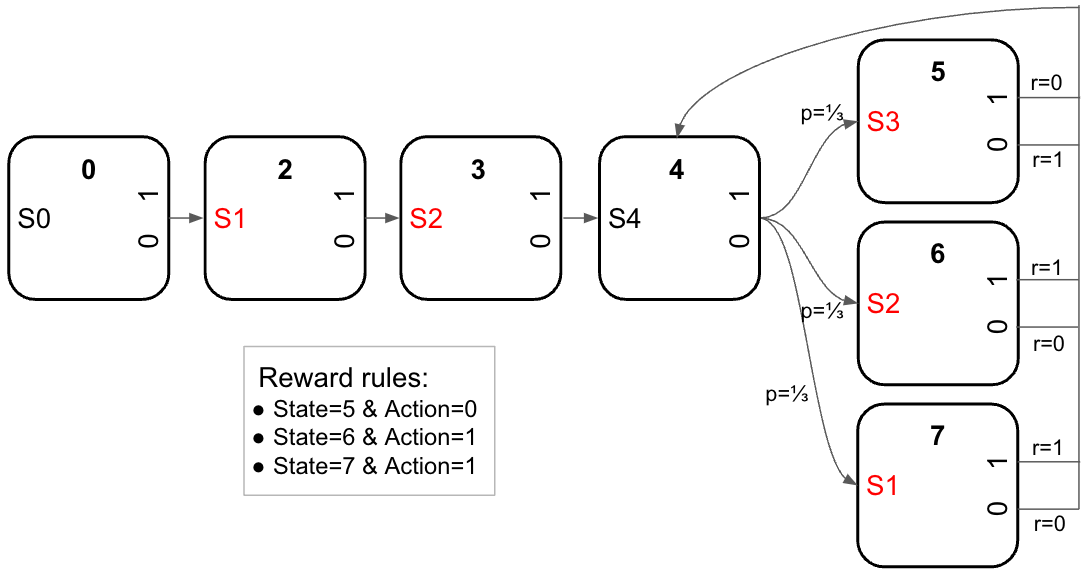}
    \caption{Familiarity detection meta-task. The agent must repeatedly determine whether incoming ``probe'' stimuli are among the two ``cue'' stimuli presented during the initial phase, or are a different stimulus.}
    \label{fig:familiarity}
\end{figure*}

In addition, we describe  a simple familiarity detection meta-task. For each task, the agent sees two ``cue''  stimuli in turn,  then is repeatedly exposed to a ``probe'' stimulus, which  may be  either one of the two previously seen cue stimuli, or another one. The agent must simply reply ``1'' if  the probe stimulus is one of the  two cue stimuli, or ''0'' otherwise. The  three stimuli are randomly generated for each new task instance. Note that this task is a simplified version of the well-known ``Omniglot'' task, in which the agent must associate arbitrary labels to a small set of novel stimuli and retrieve the correct label for presentation of an additional stimulus belonging to one of the previously seen classes. The difference is that this simplified task tests memorization of shown stimuli by familiarity detection rather than explicit label recall. This task is also easily described in our formalism (see Figure \ref{fig:familiarity}).

\section{Machine-readable description of a randomly generated meta-task}

\label{sec:json}

In Figure \ref{fig:json} we provide the JSON specification for the meta-task described in Figure \ref{fig:task3}. The JSON structure describes the transition probability table, the range of each special-state variable (that is, for each special state variable, the possible values it may take), the flag conditions (as a list of rules), the reward conditions, and the stimulus associated with each of the states. 

The transition table is a matrix of size $N_s \times N_a \times N_s$, where $N_s$ is the number of states and $N_a$ the number of actions (here $N_s=4$ and $N_a=2$). For each current state, it indicates the probability of ending in each possible state, for each of the actions that can be taken. Values of the form $100+k$ denote the $k$-th state variable; $1000+k$ denote the $k$-th probability variable; and $10000+k$ denote the $k$-th stimulus variable. Each of these is to be replaced with an actual value from the appropriate domain for each new individual task sampled from the meta-task specification.

Note that the length of the stimuli list indicates the number of states in the meta-task, and similarly the length of the special-state variable range. Similarly, the length of the state-variable ranges list indicates the number of special-state variables. 

\begin{figure}
    \centering
\begin{verbatim}
{
"T": [[[0.0, 0.5, 0.5, 0.0], [0.0, 0.0, 1.0, 0.0]], 
[[0.0, 0.5, 0.5, 0.0], [0.0, 0.0, 0.0, 1.0]], 
[[0.33333, 0.33333, 0.0, 0.33333], [0.33333, 0.33333, 0.0, 0.33333]], 
[[0.0, 1.0, 0.0, 0.0], [0.0, 1.0, 0.0, 0.0]]],
"statevariableranges": [[1, 3, 2]]
"flagconditions": [[100, -1, -1, 1]],
"rewardconditions": [[2, -1, -1, 1.0, 1.0, 1]],
"stimuli": [10000, -1, 10001, 1],
}
\end{verbatim}

    \caption{The JSON description automatically generated for the meta-task described graphically in Figure \ref{fig:task3}. T is the state transition matrix. This description fully specifies the meta-task, allowing for automated sampling. See text for details.}
    \label{fig:json}
\end{figure}

\end{document}